\documentclass[11pt,a4paper]{article}

\usepackage[utf8]{inputenc}
\usepackage[T1]{fontenc}
\usepackage{lmodern}

\usepackage[
    a4paper,
    left=2.5cm,
    right=2.5cm,
    top=2.5cm,
    bottom=2.5cm
]{geometry}

\usepackage{microtype}
\usepackage{setspace}
\usepackage{amsmath}
\usepackage{amssymb}
\usepackage{bm}

\usepackage{graphicx}
\usepackage{booktabs}
\usepackage{multirow}
\usepackage{array}
\usepackage{longtable}
\usepackage{pdflscape}
\usepackage{float}

\usepackage[numbers,sort&compress]{natbib}

\usepackage[
    colorlinks=true,
    linkcolor=blue,
    citecolor=blue,
    urlcolor=blue
]{hyperref}

\usepackage{titlesec}

\titleformat{\section}
{\large\bfseries}
{\thesection.}{0.5em}{}

\titleformat{\subsection}
{\normalsize\bfseries}
{\thesubsection.}{0.5em}{}

\DeclareUnicodeCharacter{2013}{--}
\DeclareUnicodeCharacter{2014}{---}
\DeclareUnicodeCharacter{2212}{-}
\DeclareUnicodeCharacter{00D7}{\ensuremath{\times}}
\DeclareUnicodeCharacter{00A0}{ }

\usepackage{authblk}

\title{
\textbf{A Voxel-Spacing-Aware Extension of PyRadiomics for Anisotropic Texture Analysis
}
}

\author[1]{David Corral Fontecha}
\author[2,3,4,5]{Juan Miranda Bautista}
\author[2,3,4,5]{Pablo Menendez Fernández-Miranda}
\author[6]{Andrea Trapote Fernandez}
\author[7]{Lara Lloret Iglesias}
\author[5,8]{Jose A. Vega}

\affil[1]{Department of Radiology, Complejo Asistencial Universitario de León, León, Spain}
\affil[2]{Department of Radiology, Hospital Universitario Rey Juan Carlos, Móstoles, Madrid, Spain}
\affil[3]{Health Research Institute of the Jiménez Díaz Foundation, IIS-FJD, Madrid, Spain}
\affil[4]{Department of Physical Therapy, Occupational Therapy, Rehabilitation and Physical Medicine, Rey Juan Carlos University, Madrid, Spain}
\affil[5]{Department of Morphology and Cell Biology, Grupo SINPOS, Universidad de Oviedo, Oviedo, Spain}
\affil[6]{Universidad de Oviedo, Oviedo, Spain}
\affil[7]{Advanced Computing and e-Science Group, IFCA-CSIC, Santander, Spain}
\affil[8]{Facultad de Ciencias de la Salud, Universidad Autónoma de Chile, Providencia-Santiago, Chile}

\date{}

\begin{document}

\maketitle
\begin{abstract}
\textbf{Background and Objective:}
Radiomic texture features are commonly extracted from anisotropic CT and MRI acquisitions, where identical voxel offsets may correspond to different physical distances. Standard workflows often address this issue through isotropic resampling, although interpolation modifies both image geometry and intensity values. This work aimed to implement and validate a voxel-spacing-aware extension of PyRadiomics for anisotropic medical imaging.

\textbf{Methods:}
A voxel-spacing-aware framework was integrated across the Python frontend, C wrapper layer and computational backend of PyRadiomics. The implementation preserves the discretized gray levels supplied to texture computation and introduces spacing-dependent operations only when \texttt{weightingNorm="voxelSpacing"} is enabled. GLCM uses anisotropy-relative feature-level angular aggregation. GLRLM, GLDM and GLSZM are computed on a finite-volume zero-order-hold
representation derived from the native anisotropic grid, preserving gray-level values while expressing run length, dependence size and zone size on an isotropic physical lattice. NGTDM incorporates anisotropy-relative weighted neighborhood averaging. Synthetic images with isotropic and anisotropic spacing were used for software validation.

\textbf{Results:}
The modified implementation reproduced standard PyRadiomics exactly when voxel-spacing-aware mode was disabled and remained equivalent under isotropic spacing across 75 texture features. Under anisotropic spacing, voxel-spacing-aware extraction selectively modified texture families and behaved as a distinct alternative to nearest-neighbor, linear and B-spline resampling without generating interpolated gray levels. Computational profiling showed moderate runtime and memory increases across the tested finite-volume expansion factors. Sensitivity analyses quantified the effect of finite-volume rounding and showed that spacing-aware differences persisted across binWidth settings.

\textbf{Conclusions:}
The proposed framework provides a backward-compatible implementation for incorporating voxel-spacing information into selected radiomic texture operations without generating interpolated gray levels. It establishes a technical basis for future evaluation of spacing-aware radiomics in heterogeneous clinical imaging datasets.
\end{abstract}


\section{Introduction}

Radiomics aims to transform medical images into high-dimensional quantitative descriptors that capture tissue phenotype, spatial heterogeneity and structural organization \cite{lambin2012radiomics,aerts2014decoding,mayerhoefer2020introduction}. Over the past decade, radiomic features have been investigated across numerous clinical applications, including oncology, treatment response assessment and prognostic modelling \cite{lambin2012radiomics,aerts2014decoding}. As the field has matured, increasing attention has been directed toward the standardization, reproducibility and transparent implementation of radiomic biomarkers, leading to initiatives such as the Image Biomarker Standardisation Initiative (IBSI) and the adoption of open-source software frameworks \cite{zwanenburg2020image,welch2019vulnerabilities}.

Among available radiomics platforms, PyRadiomics has become one of the most widely used open-source implementations for feature extraction from medical images \cite{van2017computational}. The library provides a standardized framework for the computation of first-order, shape and texture descriptors and is explicitly recommended as a reference implementation in many radiomics workflows. Most texture features implemented in PyRadiomics are derived from matrix-based representations of spatial relationships between voxels, following classical texture analysis formulations originally introduced by Haralick and subsequently extended within the radiomics literature \cite{haralick1973textural}. These include gray-level co-occurrence matrices (GLCM), gray-level run-length matrices (GLRLM), gray-level dependence matrices (GLDM), neighborhood gray-tone difference matrices (NGTDM) and gray-level size-zone matrices (GLSZM).

A fundamental characteristic of many clinical CT and MRI acquisitions is voxel anisotropy. While in-plane voxel spacing is often submillimetric, through-plane spacing may be substantially larger due to acquisition constraints, reconstruction settings, or clinical workflow considerations. Consequently, a displacement of one voxel along different image axes may correspond to markedly different physical distances. Despite this geometric disparity, texture matrices are commonly constructed using offsets defined in discrete index space, where neighboring voxels separated by identical index displacements are treated equivalently regardless of their physical separation.

The most common strategy for addressing anisotropy is isotropic resampling prior to feature extraction. Although resampling improves geometric consistency, it simultaneously modifies image intensities through interpolation. As a result, geometric correction becomes intrinsically coupled to signal modification, making it difficult to distinguish the effects of spatial harmonization from those introduced by interpolation itself. Previous studies have demonstrated that voxel size, image acquisition parameters and scanner-dependent factors can substantially influence radiomic feature values and reproducibility \cite{shafiq2017intrinsic,mackin2015measuring,welch2019vulnerabilities}.

An alternative approach is to avoid intensity-interpolating resampling
and incorporate physical voxel geometry directly into texture computation. However, spacing-aware formulations are not routinely implemented in standard radiomics workflows and most texture computations remain fundamentally based on index-space relationships. Moreover, not all radiomic feature families admit physically meaningful spacing corrections, raising important mathematical and software-engineering questions regarding how anisotropy should be handled within standardized radiomics software.

To the best of our knowledge, publicly available descriptions of voxel-spacing-aware radiomics implementations remain limited and detailed reports of complete frontend-to-backend voxel-spacing-aware implementations within established radiomics frameworks remain scarce.

In this work, we present a voxel-spacing-aware extension of PyRadiomics implemented across the Python frontend, C wrapper layer and computational backend of the library. Rather than generating interpolated gray levels through isotropic
resampling, the proposed framework incorporates physical voxel spacing through feature-family-specific operations while preserving the discretized gray levels supplied to texture computation. We describe the mathematical rationale underlying the proposed modifications, detail the software architecture required for their implementation, discuss compatibility with existing PyRadiomics and IBSI workflows and critically evaluate the suitability of spacing-aware formulations across different radiomic feature families. This article focuses exclusively on the software architecture, mathematical formulation and implementation details of the proposed framework. Its clinical evaluation is reported separately in an independent application study \cite{corralfontecha2026physically}.

\section{Methods}

\subsection{Geometric implications of voxel anisotropy}

Texture descriptors rely on spatial relationships between neighboring voxels. In anisotropic CT and MRI acquisitions, identical index-space displacements may correspond to different physical distances, particularly when through-plane spacing differs from in-plane spacing.

Let the voxel spacing vector and discrete offset vector be defined as

\[
\mathbf{s}=(\Delta_z,\Delta_y,\Delta_x),
\qquad
\mathbf{a}=(a_z,a_y,a_x),
\]

where \(\mathbf{a}\) represents an offset in index space. The corresponding physical distance is

\begin{equation}
d_{\mathbf{a}}^{phys}
=
\sqrt{
(a_z\Delta_z)^2+
(a_y\Delta_y)^2+
(a_x\Delta_x)^2
}.
\label{eq:phys_distance}
\end{equation}

To ensure that weighting reflects anisotropy rather than absolute directional length, the proposed implementation uses the ratio between physical and index-space distances. The index-space distance is

\begin{equation}
d_{\mathbf{a}}^{index}
=
\sqrt{a_z^2+a_y^2+a_x^2},
\label{eq:index_distance}
\end{equation}

and the relative anisotropy factor is

\begin{equation}
r_{\mathbf{a}}
=
\frac{d_{\mathbf{a}}^{phys}}
{d_{\mathbf{a}}^{index}}.
\label{eq:relative_anisotropy}
\end{equation}

For valid non-zero offsets and positive physical voxel spacings,
\(r_{\mathbf{a}} > 0\). In the GLCM implementation, unnormalized
directional weights are computed as

\begin{equation}
\widetilde{w}_{\mathbf{a}}
=
\frac{1}
{\max\left(r_{\mathbf{a}},10^{-5}\right)},
\label{eq:weight}
\end{equation}

and are subsequently normalized across the set of valid directions.
The lower bound of \(10^{-5}\) is included solely as a numerical
safeguard and does not materially affect weights for valid offsets and
positive physical voxel spacings.

Normalized weights are then obtained as

\begin{equation}
W_{\mathbf{a}}
=
\frac{\widetilde{w}_{\mathbf{a}}}
{\sum_{\mathbf{b}\in A}\widetilde{w}_{\mathbf{b}}},
\label{eq:normweight}
\end{equation}

where \(A\) denotes the set of valid offsets for a given texture descriptor. Under isotropic spacing, \(r_{\mathbf{a}}\) is constant across directions and the voxel-spacing-aware branch is defined to reproduce standard PyRadiomics behavior. The implementation is activated only through \texttt{weightingNorm="voxelSpacing"}.

\subsection{Array, spacing and orientation conventions}

PyRadiomics internally represents SimpleITK images as NumPy arrays in
\((z,y,x)\) order, whereas SimpleITK reports voxel spacing in
\((x,y,z)\) order. In the voxel-spacing-aware implementation, the
SimpleITK spacing vector is therefore reversed using
\texttt{GetSpacing()[::-1]} before spacing-dependent calculations, so
that

\[
\mathbf{s}
=
(\Delta_z,\Delta_y,\Delta_x)
\]

is aligned directly with the NumPy array axes. The corresponding spacing component is then associated with the same array axis during physical-distance calculation and finite-volume expansion.

The voxel-spacing-aware operations use voxel spacing and array-axis geometry but do not incorporate the SimpleITK image direction matrix. Consequently, the current implementation assumes that the array-axis representation and associated spacing components provide the relevant orthogonal acquisition geometry. Image-direction handling is therefore
outside the scope of the present implementation and should be considered when applying the method to data requiring explicit orientation-aware physical coordinates.

\subsection{Finite-volume representation for selected texture matrices}

For GLRLM, GLDM and GLSZM, voxel-spacing-aware extraction was implemented using a finite-volume, zero-order-hold representation of the native image grid. Let

\[
h=\min(\Delta_z,\Delta_y,\Delta_x)
\]

denote the finest voxel spacing. For each axis \(k\in\{z,y,x\}\), the finite-volume expansion factor is

\begin{equation}
q_k=
\mathrm{round}
\left(
\frac{\Delta_k}{h}
\right),
\qquad k\in\{z,y,x\}.
\label{eq:finite_volume_factor}
\end{equation}

The spacing represented by the finite-volume lattice along each axis is

\begin{equation}
\Delta_k^{*}=q_k h,
\label{eq:represented_spacing}
\end{equation}

and the corresponding relative representation error is

\begin{equation}
e_k=
\frac{\Delta_k^{*}-\Delta_k}
{\Delta_k}.
\label{eq:represented_spacing_error}
\end{equation}

The total finite-volume expansion factor is

\begin{equation}
R=q_zq_yq_x.
\label{eq:total_expansion_factor}
\end{equation}

For an input array of dimensions \(N_z\times N_y\times N_x\), the
expanded lattice contains

\begin{equation}
N_{\mathrm{expanded}}
=
(N_zq_z)(N_yq_y)(N_xq_x)
\label{eq:expanded_size}
\end{equation}

subcells.

The image and mask are expanded by repeating each voxel along axis \(k\) by \(q_k\) subcells. This operation is an exact zero-order-hold replication implemented with \texttt{numpy.repeat}. It does not generate interpolated gray levels or
mix neighboring voxel values; each subcell inherits the discretized gray level of its parent voxel. However, the operation intentionally
changes the computational lattice by replacing each native anisotropic voxel with a block of isotropic subcells. Matrix construction is then performed using the standard PyRadiomics backend on the expanded lattice. Consequently, run length, dependence size and zone size remain integer-valued matrix axes, but are expressed in isotropic physical subcells rather than anisotropic native index voxels.

Before finite-volume expansion, the implementation computes the repeat factors and expanded array shape. The helper function applies a \texttt{maximumExpansionFactor} limit of 1000 to the total repeat-factor product

\[
R=q_zq_yq_x.
\]

This safeguard limits the geometric expansion factor rather than the absolute number of expanded voxels. The repeat factors, represented spacing, relative spacing error, expanded shape, and expansion factor are reported through the logging system. Warnings are emitted when the maximum absolute relative spacing error exceeds \(10\%\) or when \(R>8\). The library implementation does not currently impose an independent limit based on the total expanded voxel count.

If spacing is isotropic, no expansion is applied and the standard PyRadiomics path is preserved. For the finite-volume GLRLM, GLDM and GLSZM paths, the discretized image and binary ROI mask are expanded using identical repeat factors along each axis. Expansion is implemented using exact array replication (\texttt{numpy.repeat}); therefore, no interpolated image values or partial-volume mask values are introduced. Each expanded mask subcell inherits the binary label of its parent voxel, producing a block-wise representation of the original ROI boundary on the expanded lattice.

For GLCM and NGTDM, computations remain on the native array. Candidate neighbors lying outside the image or outside the ROI are excluded from matrix or neighborhood accumulation. In GLCM, empty directional matrices are removed and angular weights are renormalized over the remaining valid directions. In spacing-aware NGTDM, neighborhood
normalization is performed using only the weights of valid neighbors.

\begin{figure}[H]
\centering
\includegraphics[width=\textwidth]{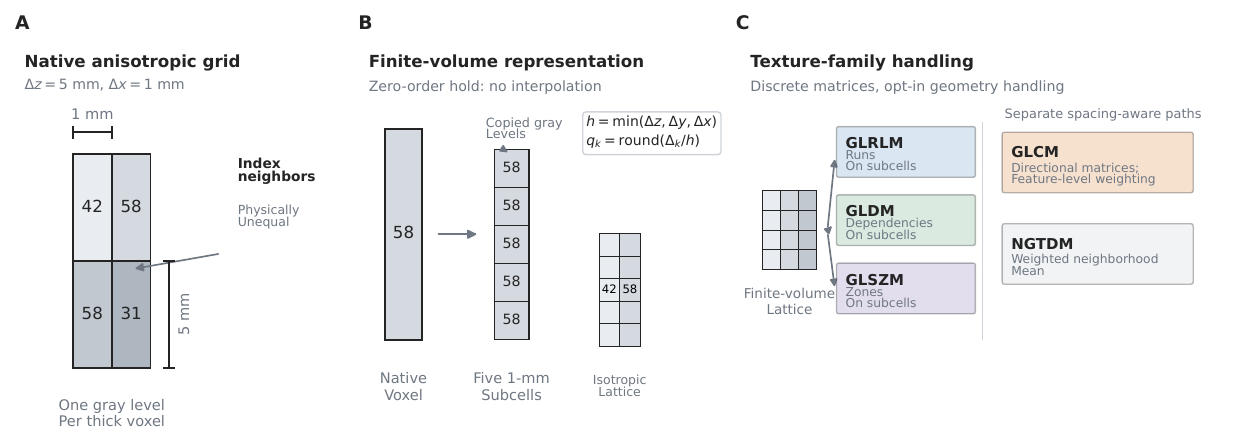}
\caption{Voxel-spacing-aware texture extraction strategy. GLCM and NGTDM
operate on the native computational lattice using anisotropy-relative
weighting, whereas GLRLM, GLDM and GLSZM are computed on a finite-volume
zero-order-hold representation. In the latter families, each anisotropic
voxel is represented by repeated isotropic subcells that inherit the
discretized gray level of the parent voxel. This operation introduces no
interpolated gray levels but changes the computational lattice on which
run, dependence and zone relationships are evaluated.}
\label{fig:finite_volume}
\end{figure}

This representation is illustrated schematically in Figure~\ref{fig:finite_volume}.

\subsection{Software implementation}

The voxel-spacing-aware framework was implemented across three software layers within PyRadiomics \cite{van2017computational}: (i) Python feature classes and configuration management, (ii) C extension wrapper functions (\texttt{\_cmatrices.c}) and (iii) computational matrix backends (\texttt{cmatrices.c}).

When \texttt{weightingNorm="voxelSpacing"} is enabled, voxel spacing information is propagated from the Python frontend through the wrapper layer to the computational backend. Physical distances and anisotropy-relative weights are incorporated according to feature-family structure. For GLCM, directional co-occurrence matrices remain separate during feature calculation, and spacing-aware weights are applied only during the final feature-level angular aggregation. NGTDM incorporates weighted neighborhood averaging, whereas GLRLM,
GLDM and GLSZM are computed on a finite-volume zero-order-hold representation derived from the native anisotropic grid.

The implementation was designed to preserve backward compatibility. When voxel-spacing-aware mode is disabled, the standard PyRadiomics execution paths, matrix definitions and feature calculations remain unchanged. The default mode is therefore intended to preserve the numerical behavior of the reference PyRadiomics implementation. In contrast, enabling \texttt{weightingNorm="voxelSpacing"} activates explicitly defined methodological extensions that incorporate voxel geometry into selected texture operations. Results obtained in this mode should not be interpreted as conventional IBSI reference values, but as outputs of the proposed voxel-spacing-aware extension.

\subsection{Feature-family-specific modifications}

Spacing-aware behavior was implemented selectively according to feature-family structure. Modifications were introduced only when voxel spacing could be incorporated without replacing the underlying descriptor semantics.

GLCM is based on directional co-occurrence statistics and therefore admits anisotropy-aware angular aggregation without altering matrix construction. In contrast, run-, dependence- and zone-based descriptors rely on discrete size measures. The feature-family-specific implementation strategies are summarized in
Table~\ref{tab:spacing_strategy}.

\subsubsection{Gray-Level Co-occurrence Matrix (GLCM)}

GLCM co-occurrence pairs are enumerated using the standard PyRadiomics
implementation. Matrix construction, discretization and Haralick
feature definitions remain unchanged. Let
\(P_{\mathbf{a}}(i,j)\) denote the directional co-occurrence matrix
associated with offset \(\mathbf{a}\).

In the proposed implementation, voxel spacing is incorporated only
during the final angular aggregation. A feature value is first computed
independently for each directional matrix, yielding
\(F(P_{\mathbf{a}})\). The final value is then obtained as

\begin{equation}
F
=
\sum_{\mathbf{a}\in A}
W_{\mathbf{a}}F(P_{\mathbf{a}}),
\label{eq:glcm_feature_aggregation}
\end{equation}

where \(W_{\mathbf{a}}\) denotes the normalized
anisotropy-relative weight defined in
Equation~\ref{eq:normweight}. Directional GLCMs are not combined before
feature computation.

Unlike matrix-level aggregation, this formulation preserves the native directional co-occurrence distributions and applies spacing-aware weighting only after feature computation. Matrix-level weighting was intentionally avoided because many Haralick descriptors are nonlinear functions of the co-occurrence matrix. Weighting matrices before feature computation would therefore modify descriptor definitions and potentially alter their mathematical interpretation. Feature-level aggregation was selected as a conservative strategy that preserves directional co-occurrence statistics while incorporating anisotropy information only during the final aggregation step. Consequently, the implementation should be interpreted as an anisotropy-aware angular aggregation strategy rather than a modification of the underlying GLCM definition.

\subsubsection{Gray-Level Run-Length Matrix (GLRLM)}

GLRLM was computed using the standard PyRadiomics backend on the finite-volume representation. Run lengths remain integer counts of contiguous elements along discrete directions, preserving the original run-length formulation \cite{galloway1975texture}. In voxel-spacing-aware mode, these counts represent isotropic subcells rather than native anisotropic voxels.

\subsubsection{Neighborhood Gray-Tone Difference Matrix (NGTDM)}

NGTDM, originally introduced by Amadasun and King
\cite{amadasun1989textural} and later standardized by IBSI
\cite{zwanenburg2020image}, depends explicitly on local neighborhood
averaging.

In the NGTDM computational backend, the relative anisotropy factor is
evaluated as

\begin{equation}
r_{\mathbf{a}}
=
\frac{d_{\mathbf{a}}^{\mathrm{phys}}}
{d_{\mathbf{a}}^{\mathrm{index}}+10^{-12}},
\label{eq:ngtdm_relative_anisotropy}
\end{equation}

and the corresponding unnormalized neighborhood weight is

\begin{equation}
\widetilde{w}_{\mathbf{a}}
=
\frac{1}{r_{\mathbf{a}}+10^{-6}}.
\label{eq:ngtdm_weight}
\end{equation}

The constants \(10^{-12}\) and \(10^{-6}\) are numerical safeguards
against division by zero and have a negligible effect for valid
non-zero offsets and positive physical voxel spacings. The resulting
weights are normalized over the valid neighborhood offsets.

The standard unweighted neighborhood mean is then replaced by

\begin{equation}
\bar I(x)
=
\frac{
\sum_{\mathbf{a}\in N(x)}
I(x+\mathbf{a})\widetilde{w}_{\mathbf{a}}
}{
\sum_{\mathbf{a}\in N(x)}
\widetilde{w}_{\mathbf{a}}
}.
\label{eq:ngtdm_weighted_mean}
\end{equation}

Consequently, neighbors located at larger physical distances contribute
proportionally less to the local intensity estimate while preserving
the original neighborhood definition. The usual NGTDM accumulation is
then performed using

\[
\left|I(x)-\bar I(x)\right|.
\]

Thus, spacing information is incorporated directly into descriptor
construction.

\subsubsection{Gray-Level Dependence Matrix (GLDM)}

GLDM was computed using the standard dependence criterion on the finite-volume representation. Dependence size remains an integer count of neighboring subcells satisfying

\[
|I(x)-I(x+a)| \le \alpha.
\]

This avoids introducing fractional dependence values while allowing anisotropic voxel geometry to affect local dependence estimation through the finite-volume lattice.

\subsubsection{Gray-Level Size-Zone Matrix (GLSZM)}

GLSZM was computed using the standard connected-zone algorithm on the finite-volume representation. Zone size remains an integer count, but in voxel-spacing-aware mode it counts isotropic subcells rather than native anisotropic voxels. Thus, topology and connectivity rules remain unchanged, while zone-size-related descriptors become sensitive to anisotropic voxel volume.

\subsubsection{Shape and first-order features}

No additional spacing-aware modifications were required for shape features because voxel spacing is already incorporated into geometric calculations. First-order descriptors are largely independent of spatial voxel relationships, except for volume-dependent quantities such as Total Energy.

\begin{table}[H]
\centering
\small
\begin{tabular}{p{0.18\textwidth}p{0.22\textwidth}p{0.24\textwidth}p{0.26\textwidth}}
\toprule
Feature family & Modification level & Use of spacing & Interpretation \\
\midrule
GLCM & Feature-level angular aggregation & Anisotropy-relative weights & Directional GLCMs are preserved and weighted after feature computation. \\
GLRLM & Finite-volume matrix construction & Zero-order-hold isotropic subcells & Run lengths remain integer counts on the expanded physical lattice. \\
NGTDM & Descriptor computation & Anisotropy-relative weighted neighborhood mean & Neighborhood averaging becomes spacing-aware. \\
GLDM & Finite-volume matrix construction & Zero-order-hold isotropic subcells & Dependence size remains integer-valued on the expanded physical lattice. \\
GLSZM & Finite-volume matrix construction & Zero-order-hold isotropic subcells & Standard connected-zone rules are applied while zone size counts isotropic subcells. \\
Shape & No new modification & Already supported & PyRadiomics already uses voxel spacing. \\
First-order & Minimal & Voxel volume where relevant & Mostly intensity-distribution based. \\
\bottomrule
\end{tabular}
\caption{Summary of spacing-aware implementation strategy by radiomic feature family.}
\label{tab:spacing_strategy}
\end{table}

\subsection{Filtered image types}

Filtered image types were largely retained from standard PyRadiomics. Wavelet, gradient and intensity-transformation filters were not functionally modified; the gradient filter already uses image spacing by default through \texttt{gradientUseSpacing=True}. Limited filter-level adjustments included spacing-aware LoG handling and conservative disabling of anisotropic 3D LBP generation when no physically justified extension was available.

\subsection{Regression and compatibility testing}

Software verification included both the existing upstream PyRadiomics
regression suite and dedicated voxel-spacing-aware tests. In the evaluated development overlay, 3568 upstream tests passed, including reference feature and matrix tests designed to detect changes in
standard PyRadiomics behavior.

Additional spacing-aware tests verified isotropic equivalence by comparing default extraction with
\texttt{weightingNorm="voxelSpacing"} under isotropic voxel spacing. The current test suite does not contain an independent IBSI benchmark for the new spacing-aware mode. It also does not yet provide dedicated anisotropic reference-value fixtures for every spacing-aware feature family. The regression results therefore establish backward
compatibility with the tested PyRadiomics implementation rather than formal IBSI conformance of the proposed extension.

\subsection{Synthetic software validation}

A synthetic software-validation framework was designed to verify the correctness, behavior and computational characteristics of the proposed voxel-spacing-aware implementation under controlled conditions. All experiments were performed using deterministic three-dimensional synthetic phantoms and were intended exclusively for software validation rather than clinical performance assessment.

The primary validation experiment assessed backward compatibility, isotropic equivalence and anisotropic activation. Backward compatibility was defined as exact numerical agreement between the original PyRadiomics implementation and the modified implementation with voxel-spacing-aware functionality disabled. This test was intended to verify that the new code did not alter the established default PyRadiomics behavior. Evaluation of the voxel-spacing-aware mode was treated separately because its outputs represent newly defined methodological extensions rather than conventional IBSI reference values.
A three-dimensional image containing predefined intensity patterns and a binary region-of-interest mask was generated using both isotropic and anisotropic spacing configurations. Radiomic features were extracted using (i) the original PyRadiomics implementation, (ii) the modified implementation with voxel-spacing-aware functionality disabled and (iii) the modified implementation with \texttt{weightingNorm="voxelSpacing"} enabled. Feature extraction was restricted to GLCM, GLRLM, GLDM, NGTDM and GLSZM, while shape features were excluded because physical spacing is already incorporated in standard PyRadiomics geometric calculations. All texture features were computed using fully three-dimensional (3D) matrix construction with \texttt{force2D=False}. No slice-wise 2D or 2.5D texture extraction was used in the present validation experiments.

To compare the proposed framework with conventional preprocessing strategies, additional anisotropic synthetic phantoms were generated with through-plane spacing ratios of 2:1:1, 3:1:1 and 5:1:1 in array order ((z,y,x)). Each phantom contained a binary ROI, an in-plane intensity gradient, through-plane alternating bands, local heterogeneous texture and fixed-seed noise. Texture features were extracted using native PyRadiomics without resampling, isotropic resampling with nearest-neighbor, linear or B-spline interpolation, and the voxel-spacing-aware implementation. Mask resampling was always performed using nearest-neighbor interpolation.

For each resampling comparison, a feature-level output table was generated containing the native value, comparison value, absolute difference, relative difference, and extraction method. Results were summarized across individual feature--phantom comparisons and by texture family.

Computational profiling was performed using synthetic volumes of \(16\times64\times64\), \(32\times96\times96\), and \(48\times128\times128\) voxels. The corresponding native ROI sizes were 12,126, 50,328, and 139,028 voxels. Expansion factors of \(R=1,2,3,5\), and 7 were evaluated. Runtime and approximate peak memory consumption were recorded together with the repeat factors \(q_k\), represented spacings \(\Delta_k^{*}\), relative representation errors \(e_k\), expanded array dimensions, and expanded ROI sizes.

Sensitivity to non-integer spacing ratios was evaluated using through-plane spacings of 2.0, 2.5, 3.0, 3.5, 5.0, and 5.5 mm with fixed 1 mm in-plane spacing. Finally, discretization sensitivity was assessed by repeating feature extraction with \texttt{binWidth} values of 5, 10, 25, and 50.

\section{Results}

\subsection{Implementation outcome}

The voxel-spacing-aware framework was integrated across the Python frontend, shared configuration and spacing utilities, the C extension wrapper, and the computational backend of PyRadiomics. The
implementation introduced an optional execution mode through \texttt{weightingNorm="voxelSpacing"} without altering the standard external feature-extraction workflow.

The evaluated development overlay passed all 3568 tests executed from the upstream PyRadiomics regression suite. Dedicated synthetic tests
also confirmed exact isotropic equivalence between default extraction and the spacing-aware execution mode under isotropic voxel spacing.

Spacing-aware functionality was implemented selectively according to feature-family structure. GLCM used anisotropy-relative feature-level angular aggregation. GLRLM, GLDM and GLSZM were computed on a finite-volume zero-order-hold
representation derived from the native anisotropic grid. NGTDM incorporated anisotropy-relative weighted neighborhood averaging.

\subsection{Software validation}

The modified implementation reproduced original PyRadiomics results exactly when voxel-spacing-aware mode was disabled. For the anisotropic synthetic image, all 75 evaluated texture features were identical between the original and modified implementations. Similarly, under isotropic voxel spacing, enabling \texttt{weightingNorm="voxelSpacing"} produced identical results to the default implementation, with mean and maximum absolute differences equal to zero.

Under anisotropic voxel spacing, activation of
\texttt{weightingNorm="voxelSpacing"} modified all evaluated GLCM,
GLRLM and NGTDM features, 11 of 14 GLDM features and 8 of 16 GLSZM
features. Median relative differences were 10.1\% for GLCM, 26.1\%
for GLRLM, 33.6\% for NGTDM, 49.1\% for GLDM and 40.0\% for GLSZM.

Additional validation showed that voxel-spacing-aware extraction was not equivalent to conventional isotropic resampling. Across the three anisotropic synthetic phantoms, corresponding to 225 feature--phantom comparisons, nearest-neighbor resampling changed 192/225 comparisons (85.3\%), whereas linear and B-spline resampling changed 225/225 comparisons (100\%). Voxel-spacing-aware extraction changed 192/225 comparisons (85.3\%).

The median of the family-level median relative differences was 0.333 for nearest-neighbor resampling, 0.304 for linear resampling, 0.302 for B-spline resampling, and 0.307 for voxel-spacing-aware extraction. The minimum method-level Spearman correlations were 0.900 for nearest-neighbor and voxel-spacing-aware extraction and 0.800 for linear and B-spline resampling. These results confirm that the proposed method is not numerically equivalent to resampling, despite producing differences of comparable magnitude, because it preserves the gray levels supplied to texture computation rather than generating interpolated gray levels.

Computational profiling showed increasing memory requirements with finite-volume expansion while runtime remained moderate across the tested synthetic settings. The most demanding completed experiment used a \(48\times128\times128\) volume with spacing \(0.7{:}0.7{:}5.0\) mm, producing repeat factors \(q=(1,1,7)\), an expanded lattice of \(48\times128\times896\), and 973,196 ROI subcells. Runtime was 1.74 s and approximate peak memory consumption was 90.55 MB. Across all tested configurations, median peak memory increased from 2.83 MB at \(R=1\) to 13.56 MB at \(R=7\), while the maximum observed runtime remained below 1.75 s.

Finite-volume rounding introduced measurable geometric approximation for non-integer spacing ratios. Integer spacing ratios were represented without error. A \(2.5{:}1{:}1\) spacing ratio was represented using \(q_z=2\), corresponding to a relative through-plane representation error of \(-20.0\%\), whereas a \(3.5{:}1{:}1\) ratio was represented using \(q_z=4\), corresponding to an error of \(+14.3\%\). Across adjacent tested spacing-ratio configurations, the largest family-level median relative feature difference was 0.167 for GLSZM, and the lowest Spearman correlation was 0.900 for NGTDM. These results do not eliminate the integer-rounding approximation but quantify its effect and make the represented geometry auditable.

Binning sensitivity analysis showed that voxel-spacing-aware differences persisted across \texttt{binWidth} values. Median relative differences versus native extraction ranged from 14.5\% to 23.3\%, with Spearman correlations ranging from 0.973 to 0.982.

\begin{table}[H]
\centering
\small
\begin{tabular}{
    @{}
    p{0.25\textwidth}
    p{0.24\textwidth}
    p{0.22\textwidth}
    p{0.19\textwidth}
    @{}
}
\toprule
\textbf{Validation component}
& \textbf{Evaluated data}
& \textbf{Main metric}
& \textbf{Result} \\
\midrule

Upstream regression suite
& 3568 tests
& Passed tests
& 3568/3568 \\

Default compatibility
& 75 features
& Mean/max absolute difference
& 0 / 0 \\

Isotropic equivalence
& 75 features
& Mean/max absolute difference
& 0 / 0 \\

Anisotropic activation
& 75 features
& Median relative difference
& 10.1--49.1\% \\

Rounding sensitivity
& Seven spacing ratios
& Representation error
& \(-20.0\%\) to \(+14.3\%\) \\

Computational profiling
& Three volume sizes; \(R=1\)--7
& Maximum runtime / memory
& 1.74 s / 90.55 MB \\

Resampling comparison
& 225 feature--phantom comparisons
& Changed comparisons
& 85.3--100\% \\

Binning sensitivity
& Four \texttt{binWidth} values
& Median relative difference
& 14.5--23.3\% \\

\bottomrule
\end{tabular}

\caption{Summary of synthetic software validation, finite-volume
rounding, computational profiling, resampling comparisons, and
discretization sensitivity.}
\label{tab:synthetic_validation}
\end{table}

The principal software-validation results are summarized in Table~\ref{tab:synthetic_validation}.

\section{Discussion}

The present work introduces a voxel-spacing-aware extension of PyRadiomics that incorporates physical voxel geometry into selected texture operations while preserving the discretized gray levels supplied to texture computation. Unlike isotropic resampling, which modifies both geometry and image intensities through interpolation, the proposed framework addresses anisotropy at the feature-extraction stage without generating interpolated gray levels. Synthetic validation demonstrated three key properties: exact backward compatibility with standard PyRadiomics when spacing-aware mode is disabled, exact equivalence under isotropic spacing and selective activation under anisotropic spacing.

A central design decision was the use of relative anisotropy rather than absolute physical distance for angular weighting. Direct weighting by physical distance would alter descriptor behaviour even under isotropic spacing by systematically penalizing diagonal offsets. By weighting according to the ratio between physical and index-space distances, the proposed formulation remains equivalent to standard PyRadiomics under isotropic conditions while selectively accounting for anisotropy when physical and discrete spatial relationships diverge.

The implementation highlights that spacing-aware radiomics requires feature-family-specific treatment. For GLCM, feature-level aggregation was adopted to preserve directional co-occurrence distributions and avoid altering non-linear Haralick descriptors through matrix averaging. NGTDM incorporates anisotropy-relative weighting directly into neighborhood averaging. For GLRLM, GLDM and GLSZM, the finite-volume representation preserves standard matrix semantics by keeping run length, dependence size and zone size as integer-valued axes, while allowing these quantities to be measured on an isotropic physical lattice rather than on anisotropic native voxels.

The relationship with PyRadiomics reference behavior and IBSI should be considered separately. When \texttt{weightingNorm="voxelSpacing"} is disabled, the modified library uses the standard PyRadiomics execution paths. Backward compatibility
was additionally assessed using the upstream PyRadiomics regression suite, for which 3568 tests passed in the evaluated development overlay, together with dedicated synthetic tests of isotropic equivalence. These tests demonstrate preservation of the reference PyRadiomics behavior covered by the existing test suite, but should not be interpreted as independent IBSI conformance testing.

When \texttt{weightingNorm="voxelSpacing"} is enabled, spacing information is incorporated into selected texture operations and the resulting values represent methodological extensions rather than conventional IBSI reference values. For GLCM and NGTDM, the extension modifies angular feature aggregation or neighborhood averaging. For GLRLM, GLDM and GLSZM, standard matrix construction is applied after finite-volume zero-order-hold expansion rather than directly to the native index lattice. No independent IBSI benchmark extraction was performed for the voxel-spacing-aware mode, and the proposed outputs are therefore not claimed to be IBSI compliant or IBSI-equivalent. Dedicated anisotropic reference fixtures and independent IBSI
benchmarking remain appropriate targets for future validation.

Several limitations should be acknowledged. First, the finite-volume representation requires integer repeat factors. For non-integer spacing ratios, the represented spacing \(\Delta_k^{*}=q_kh\) may differ from the acquired spacing \(\Delta_k\). Although this approximation was quantified and is reported through \(e_k\), it remains the principal mathematical limitation of the current implementation. Future approaches could consider rational approximations, direct physical traversal, or streaming matrix construction.

Second, finite-volume expansion increases array size approximately in proportion to \(R=q_zq_yq_x\). The implemented expansion-factor limit, expanded-shape calculation, logging, and warning thresholds provide safeguards against extreme finite-volume expansion, but larger clinical volumes and highly anisotropic multi-axis acquisitions may still require tiling or streaming implementations. The computational results reported here therefore support moderate-cost operation only within the tested volume sizes and expansion factors. However, because the current library-level safeguard is based on
\(R\) rather than on the absolute expanded voxel count, memory requirements may still become substantial for large native volumes. A future implementation could additionally impose an explicit pre-allocation memory or expanded-voxel limit.

Third, the present study used deterministic synthetic phantoms designed
for software validation. It does not establish clinical superiority, test--retest reproducibility, or predictive performance. The clinical impact of the implementation has been evaluated separately in an independent application study \cite{corralfontecha2026physically}. Finally, the spacing-aware operations are opt-in methodological extensions and are not claimed to produce conventional IBSI reference values.

Overall, the proposed framework provides a practical and backward-compatible strategy for incorporating physical voxel geometry into radiomic texture computation without generating interpolated gray levels. The implementation establishes a foundation for future investigations of spacing-aware radiomics in clinical and methodological settings.

\section{Conclusion}

This work presents a voxel-spacing-aware extension of PyRadiomics that incorporates physical voxel geometry into selected texture operations while avoiding the generation of interpolated gray levels. Rather than relying exclusively on isotropic resampling, the proposed framework introduces feature-family-specific spacing-aware operations, including native-lattice weighting for GLCM and NGTDM and finite-volume zero-order-hold representation for GLRLM, GLDM and GLSZM.

The implementation was integrated across the Python frontend, C wrapper layer and computational backend of PyRadiomics while maintaining full backward compatibility through an optional execution mode. GLCM was handled through feature-level angular aggregation, whereas GLRLM, GLDM and GLSZM were computed on a finite-volume zero-order-hold lattice. Synthetic validation demonstrated exact agreement with standard PyRadiomics when spacing-aware extraction was disabled and exact equivalence under isotropic voxel spacing. Under anisotropic conditions, spacing-aware computation produced selective feature-family activation consistent with the proposed mathematical formulation.

These results establish a practical and extensible framework for incorporating voxel-spacing information into radiomic texture analysis and provide a foundation for future studies evaluating its impact on feature reproducibility, robustness and clinical modelling.

\section*{Acknowledgments}

The authors thank the developers and contributors of the PyRadiomics project for providing an open-source framework that enabled the implementation and evaluation of the proposed methodology.

\section*{Funding}

This research received no specific grant from any funding agency in the public, commercial, or not-for-profit sectors.

\section*{Declaration of Competing Interest}

The authors declare that they have no known competing financial interests or personal relationships that could have appeared to influence the work reported in this paper.

\section*{Ethics Statement}

This study did not involve human participants, patient data, animal experiments, or clinical interventions. Validation was performed exclusively using synthetic imaging data generated for software verification purposes. Therefore, ethics committee approval and informed consent were not required.

\section*{Data Availability}

The synthetic validation datasets and derived results generated during this study are available from the corresponding author upon reasonable request.

\section*{Code Availability}

The voxel-spacing-aware modifications described in this work are being prepared for public release. The implementation is intended to be submitted as a contribution to the PyRadiomics project and will be made available through a public source-code repository following review and integration procedures. Source code may also be obtained from the corresponding author upon reasonable request.

\bibliographystyle{unsrtnat}
\bibliography{bibtex}

\end{document}